# A Bayesian approach for prompt optimization in pre-trained language models.

Antonio Sabbatella[2], Andrea Ponti[1], Antonio Candelieri[1],

Ilaria Giordani[3], Francesco Archetti[2]

University Milano Bicocca, Milano, Italy
a.sabbatella@campus.unimib.it [2]
{andrea.ponti, antonio.candelieri, francesco.archetti}@ unimib.it
ilaria.giordani@ oaks.cloud

**Abstract.** A prompt is a sequence of symbol or tokens, selected from a vocabulary according to some rule, which is prepended/concatenated to a textual query. A key problem is how to select the sequence of tokens: in this paper we formulate it as a combinatorial optimization problem. The high dimensionality of the token space compounded by the length of the prompt sequence requires a very efficient solution. In this paper we propose a Bayesian optimization method, executed in a continuous embedding of the combinatorial space. In this paper we focus on hard prompt tuning (HPT) which directly searches for discrete tokens to be added to the text input without requiring access to the large language model (LLM) and can be used also when LLM is available only as a black-box. This is critically important if LLMs are made available in the Model as a Service (MaaS) manner as in GPT-4. The current manuscript is focused on the optimization of discrete prompts for classification tasks. The discrete prompts give rise to difficult combinatorial optimization problem which easily become intractable given the dimension of the token space in realistic applications. The optimization method considered in this paper is Bayesian optimization (BO) which has become the dominant approach in black-box optimization for its sample efficiency along with its modular structure and versatility. In this paper we use BoTorch, a library for Bayesian optimization research built on top of pyTorch. Albeit preliminary and obtained using a "vanilla" version of BO, the experiments on RoBERTa on six benchmarks, show a good performance across a variety of tasks and enable an analysis of the tradeoff between size of the search space, accuracy and wall clock time.

## 1 Introduction

The growing diffusion of pre-trained large language models o LLM in all domains of government, business and science has been generating an increasing interest in the developments of sw tools designed to provide a more effective adaptation of LLMs to specific tasks.

1 Department of Economics, Management, and Statistics, University of Milan-Bicocca, Milan, Italy
2 Department of Computer Science, Systems and Communications, University of Milan-Bicocca, Milan, Italy

The approach originally proposed to adapt pre-trained LLMs to a specific task was based on fine-tuning the model: albeit effective, this approach faces computational and privacy issues which hampers its practical application.

To improve this adaptation, the use of natural language prompt has become increasingly important. Let it be text or images, an appropriate prompt makes the output of the model better suited to the users' task.

A prompt is a sequence of symbol or tokens, selected from a vocabulary according to some rule, which is prepended/concatenated to a textual query.

A key problem is how to select the sequence of tokens: in this paper, we formulate it as a combinatorial optimization problem. The high dimensionality of the token space compounded by the length of the prompt sequence requires a very efficient solution. In this paper we propose a Bayesian optimization method, executed in a continuous embedding of the combinatorial space, whose solution is mapped back to the search space via discretization. Prompt-based methods can be categorized in two types.

Soft prompt tuning (SPT) which requires gradient propagation leaving other model parameters frozen (Lester et al. 2021).

Hard prompt tuning (HPT) which directly searches for discrete tokens to be added to the text input: unlike SPT methods which require access to the LLM, HPT can be used also when LLM is available only as a black-box.

This is critically important if LLMs are made available in the Model as a Service (MaaS) manner, as in GPT-4. The reference scenario in this paper is that PLMs are seen as input/output machines, without assuming to have access to their parameters but only to their output for a given input. There are some advantages in developing tools in this black-box setting. First it mitigates the security risk of the cloud infrastructure: the model parameters are hidden and known only to the service providers giving access only to the query and prediction interface. The black-box setting is also aligned with the interest of the final user allowing a simpler and more economical service than accessing the model's gradient. Other advantages are a reduction of the transmission costs and the possibility to prevent the data leakage.

The solution developed in this paper is addressed to prompt learning and optimization under the black-box constraint. The current manuscript is focused on the optimization of discrete prompts.

The discrete prompts give rise to difficult combinatorial optimization problem which easily become intractable given the dimension of the tokenizer space in realistic applications. Several papers, which will be outlined in sect. 4, have already considered optimization of discrete prompts using mostly CMA-ES as the optimization method.

The optimization method considered in this paper is Bayesian optimization, which has become the dominant approach in black-box optimization. (Archetti & Cancellieri 2019).(Garnett, R 2023). The main advantage of Bayesian optimization (BO) is its sample efficiency along with its modular structure and versatility. Compared to other black-box optimization, as evolutionary search or particle swarm optimization amongst others, BO has a significant computational overhead, because its features of versatility, flexibility and sample efficiency are induced by a more complex computational architecture. However, the implementation of BO is enabled by one of the many software frameworks available off-the-shelf from industrial and academic research groups.

In this paper, we use BoTorch, a library for Bayesian optimization research built on top of pyTorch. BoTorch provides a modular and flexible interface for composing Bayesian optimization algorithms. BoTorch bridges the gap between research and production by allowing a BO researcher to test new specific modules, e.g. acquisition functions, and a data scientist from industry to use a reliable production-grade implementation that can be seamlessly integrated l with other higher level platforms.
A specific outline of the BO algorithm used in this paper is given in the appendix B.

The contributions of this work are:

- An optimization model which works directly on the space of tokens
- A new black-box prompt optimization method
- A computational analysis of benchmarking data sets.
- An analysis of the tradeoff between size of the search space, accuracy and wall clock time.

This manuscript is organized in the following sections.

- Sect. 2 provides the formulation of the model and optimization problem.
- Sect. 3 gives the basic background about BO, an outline of the problem of structured inputs and of the BoTorch library used in this paper.
- Sect .4 provides a broad analysis of the papers on prompt optimization, focused on the black-box discrete models. (hard prompting).
- Sect. 5 establishes the experimental setting, in particular the datasets used and the algorithmic baselines.
- Sect. 6 presents the computational results.
- Sect .7 analyzes the impact of the key parameters of the experimental setting over the results.
- Sect. 8 contains conclusions, limitations, and perspectives of the proposed approach.

## 2 Formulation of the model and the optimization problem

Prompt optimization aims to find discrete tokens to be concatenated directly to the test queries with the goal of maximizing the performance on a downstream task. The objective function is given by the loss between the PLM prediction and the input labels.

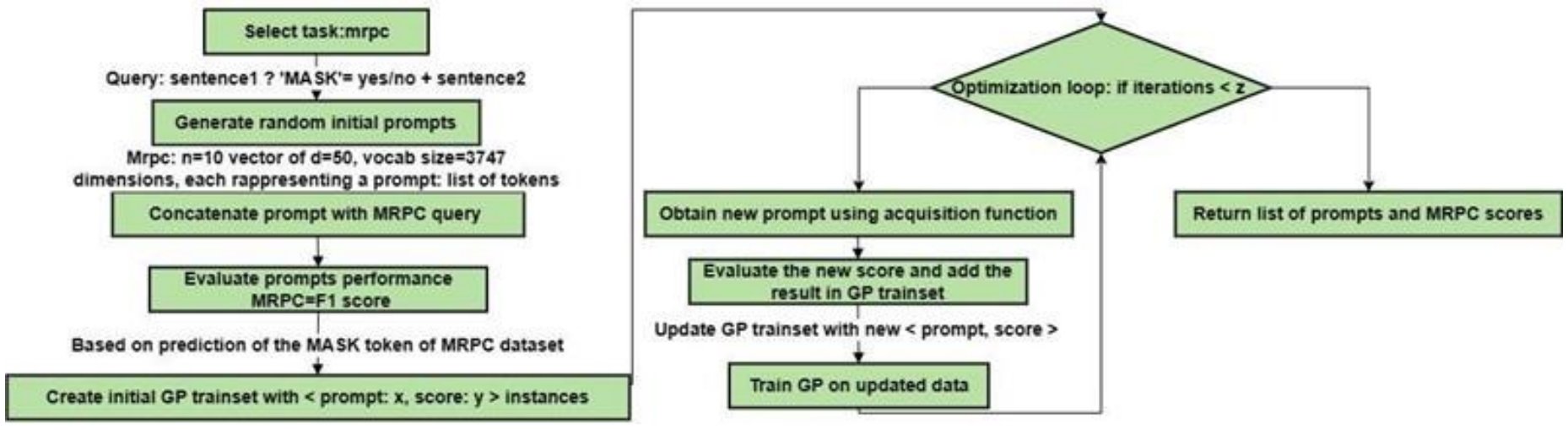

**Fig. 1.** The workflow of the proposed method

```
Algorithm Discrete Prompt Optimization

Require:
  Task T
  LLM Model M
  Dataset D
  Acquisition function A
  Objective function E
  Number of initial prompts N

Generate N initial random prompts P_1,...,P_N
for i ≤ N do
  score_i = E(M, D, P_i)
end for

GP = GaussianProcess({(P_1, score_1),...,(P_N,score_N)})

for k ≤ K do
  P_new = A(GP)
  score_new = E(M, D, P_new)
  GP.update((P_new,score_new))
end for

return GP.get_b_top_prompts(b:int)
```

**Fig. 2.** Pseudocode of the Prompt Optimization

# 3 Bayesian Optimization

## 3.1 The basic structure

Bayesian optimization (BO) is the de-facto standard for black-box optimization problems, in particular when the cost of the evaluation of the objective function and constraints is high and sample efficiency is the key requirement.

The initialization of the BO procedure is given by a set of function evaluations D = ($x_1$, $y_1$), … ($x_n$, $y_n$) with $y_i$ is a noisy observation of f($x_i$). A surrogate model, usually a Gaussian process (GP) GP, trained on D, yields a predictive model of the objective p(y | x, D).

This predictive posterior enables to build an acquisition function whose optimization yields the "best" candidates in the search space for the next function evaluations. The key role of the acquisition function is to manage the trade/off between exploitation, i.e. focused on points in the search space where the mean of p(y | x, D) is high /low (respectively for maximization and minimization problems) and exploration where the variance of p(y | x, D) is high. As new data are available the surrogate model is updated yielding more accurate predictions. This process of sequential candidate selection has a significant computational overhead, with respect to other black-box algorithms as evolutionary ones, which is more than off-set, when the function evaluation is expensive, by the sample efficiency of BO.

## 3.2 BO over structured inputs

A limitation of BO is that working over structured inputs (like graphs, categorical and integer variables) adds another layer of complexity. This comes from structuring the GP (Deshwal et al., 2021). combining latent space and structured kernels and from the optimization of the acquisition function. Several strategies have been proposed in the literature. One is to embed structured inputs into a lower dimensional continuous space where BO is executed whose output is subsequently decoded back into the structured space. (Maus et al., 2022).

Another is probabilistic reparameterization (Daulton et al. 2022) where instead of directly optimizing the acquisition function over the search space we minimize its expectation over a probability distribution defined by continuous parameters. A simple and widely used approach, which is used in this paper, is to apply a continuous relaxation of the discrete search space, solve the continuous optimization problem using BO and discretize the results.

## 4 Related works

The prompt optimization is a difficult and high-dimensional problem which has been addressed using different modeling and algorithmic strategies which will be briefly commented upon in this section. AUTOPROMPT (Shin et al., 2020) automatically generates prompts for a diverse set of tasks based on a gradient driven search.

Prompt learning is a class of methods for LLM adaptation and has become an efficient alternative to full model fine-tuning. (Liu, et al., 2021).
Earlier methods referred to as soft tuning are based on parameter -efficient fine-tuning techniques (Hu et al., 2021). (Lester et al., 2021) and directly optimize in the embedding space leaving the other model parameters frozen.

Soft prompting methods require back-propagation of gradients through the LLM. Soft prompting methods are also referred to as white model.

Hard prompt Tuning directly searches for discrete token to be added to text input, does not use internal knowledge about the pretrained LLM and only requires for the LLM to be accessible as a black box.

The approach in (Sun et al., 2022 a) optimizes a continuous prompt prepended to the input text. Instead of optimizing in the original high-dimensional prompt space the optimization is performed in a randomly generated subspace of a lower intrinsic dimensionality. Two loss functions are considered cross entropy and hinge loss and optimized using CMA-ES (Covariance Matrix Adaptation Evolution Strategy). This approach is further developed in (Sun et al., 2022 b) which used a normal distribution in the projection instead of a uniform distribution.

Another approach to black-box discrete prompt learning is proposed in (Diao et al., 2023) which applies a policy gradient policy to estimate the gradients of the parameters of the categorical distribution of each discrete prompt.

Alternative spaces for token-based optimization have been also proposed.
(Zhang et al., 2022) provide query dependent discrete prompts whose optimization is performed using reinforcement learning. and (Prasad et al., 2023) provide an automated procedure for improving prompts via an iterative local edit and gradient free search.

A different approach is proposed in (Zhou et al., 2023) based on the observation that only some tokens exert a disproportioned influence on the LLM prediction and propose to first cluster and then prune the search space to focus exclusively on influential tokens.

Another gradient free approach is proposed in (Shen et al. 2023) which adds a layer of uncertainty quantification to improve the reliability of prompt tuning and consider a strict notion of black-box setting which is likely-hood free. Borrowing from Sun et al. 2022 (b) and (Daulton et al., 2022) they also consider the lower dimensionality parametrization but instead of learning a point estimate learn a distribution again using CMA-ES.

Close to the approach proposed in the current paper a paper has recently proposed Byesian optimization for black box prompting adversarial optimization. (Maus et al., 2023) which projects the token space into an embedding space of a lower dimension. BOTorch offers several solutions for the high dimensional optimization: TuRBO (Eriksson et al. ,2019) which mitigates the curse of dimensionality limiting the search space to be within a hyper -rectangle trust region.

Other high dimensional algorithms vailable in BoTorch are BAxUS (Papenmeier et al., 2022) and SAAsBO (Eriksson et al. , 2021). Other strategies for high dimensional BO are: (Candelieri et al. , 2023 a ) and (Candelieri et al. , 2023 b) where to each function evaluation is associated a discrete probability distribution and the acquisition function is cast as a functional over discrete probability distributions embedded in a Wasserstein space.

A different strategy assuming the availability of computationally cheaper and less accurate information sources (Poloczek et. al. 2017). The sources can be obtained subsampling the training and testing datasets and augmenting the dataset on which the GP is trained also on cheaper sources. (Candelieri et al. 2021)

# 5 Experimental setting

## 5.1 Datasets and Evaluation Metrics

We include the following 6 datasets:
**MNLI** (Williams et al., 2017)**, QQP** (Sharmaet et al., 2019), **SST-2** (Semwal et al. , 2018 ).**MRPC** (Vrbanec et al., 2020 ) , **QNLI (** Demszky et al. , 2018 ) ,**RTE (**Poliak et al., 2020).

**Metrics**: MNLI: acc, QQP: F1, SST-2: acc, MRPC: F1, QNLI: acc, RTE: acc.

## 5.2 Baseline

- PromptTuning (Lester et al., 2021): a frozen RoBERTa-large model with continuous prompt embeddings prepended to the input and learned by gradients (white-box). (Lester et al., 2021).
- AutoPrompt (Shin et al., 2020): a frozen RoBERTa-large model with discrete prompts optimized based on gradient-guided search (white-box).
- Black-Box Tuning for Language-Model-as-a-Service. (Sun et al., 2022).
- P-tuning v2 (Liu et al., 2021).

# 6 Experimental results

**Table 1.** Data table from (Diao et al. 2022). The performance of BO is given in Table 1 for each task (column) and method (row). For white-box methods, the last column and row in table 1 give respectively the average scores for methods and for tasks.
For black-box methods, the last row gives the results of Bayesian optimization, averaged over three runs for each task, with the relative standard deviation given by the small subscript.

| Dataset | MNLI | QQP | SST-2 | MRPC | QNLI | RTE | avg, |
|---|---|---|---|---|---|---|---|
| White-Box Methods | | | | | | | |
| FT | $50.8_{1.2}$ | $60.8_{1.9}$ | $86.5_{2.0}$ | $78.4_{1.3}$ | $53.2_{1.8}$ | $55.6_{2.5}$ | 64,2 |
| Prompt Tuning | $36.5_{0.9}$ | $50.2_{1.5}$ | $70.7_{2.6}$ | $52.7_{3.4}$ | $53.5_{1.6}$ | $56.3_{1.6}$ | 53,3 |
| P. Tuning-v2 | $44.2_{1.7}$ | $57.4_{2.4}$ | $80.4_{1.2}$ | $62.4_{2.0}$ | $51.5_{1.3}$ | $53.1_{1.7}$ | 58,2 |
| AutoPrompt | $40.1_{1.5}$ | $45.7_{1.3}$ | $71.5_{2.1}$ | $63.8_{3.1}$ | $50.2_{1.3}$ | $52.1_{1.6}$ | 53,9 |
| Feature Probe | $46.5_{1.8}$ | $56.3_{1.1}$ | $79.5_{1.6}$ | $68.9_{1.7}$ | $50.5_{0.2}$ | $54.1_{2.5}$ | 59,3 |
| **Avg. W.B.** | 41.825 | 54.08 | 77.72 | 65.24 | 51.78 | 54.24 | 57,78 |
| Black-Box Methods | | | | | | | |
| Manual Prompt | $35.9_{1.3}$ | $49.8_{0.9}$ | $72.2_{2.1}$ | $70.4_{1.6}$ | $49.2_{1.1}$ | $48.2_{0.6}$ | 54.3 |
| ICT | $37.2_{1.6}$ | $50.1_{0.9}$ | $82.8_{2.1}$ | $72.1_{2.3}$ | $50.8_{0.5}$ | $49.3_{2.3}$ | 57.1 |
| BBT | $40.6_{2.5}$ | $55.2_{3.1}$ | $85.3_{3.9}$ | $66.4_{3.7}$ | $\mathbf{55.4}_{3.2}$ | $52.6_{2.2}$ | 59.3 |
| RLPrompt | $\mathbf{42.8}_{3.2}$ | $53.7_{2.2}$ | $\mathbf{88.4}_{1.9}$ | $68.9_{2.1}$ | $52.6_{1.4}$ | $51.8_{1.8}$ | 59.7 |
| BDPL | $42.5_{1.8}$ | $\mathbf{56.4}_{1.9}$ | $87.6_{2.1}$ | $\mathbf{78.1}_{3.7}$ | $53.1_{1.1}$ | $\mathbf{53.5}_{0.9}$ | 61.9 |
| **Avg B.B.** | 39.8 | 53.04 | 83.26 | 71.18 | 52.22 | 51.08 | 58.46 |
| **BO** | $29.6_{1.7}$ | $53.8_{0.0}$ | $86.2_{2.4}$ | $\mathbf{78.1}_{4.6}$ | $52.9_{1.3}$ | $51.0_{1.1}$ | 58.6 |

BO has the highest score for MRPC and is very close to the top scores for most of the others. The performance of BO is significantly worse than the others on MNLI. A possible explanation is that MNLI has the largest vocabulary and a more sophisticated encoding than continuous relaxation might yield a better result. Significantly, the best BO is better than the average BB and the average BO is about the same.

**Table 2.** Cardinality of the vocabulary

| MNLI | QQP | SST-2 | MRPC | QNLI | RTE |
|---|---|---|---|---|---|
| 117056 | 61571 | 3747 | 7940 | 3163 | 46992 |

**Table 3.** Prompt length

| MNLI | QQP | SST-2 | MRPC | QNLI | RTE |
|---|---|---|---|---|---|
| 10 | 25 | 50 | 50 | 50 | 50 |

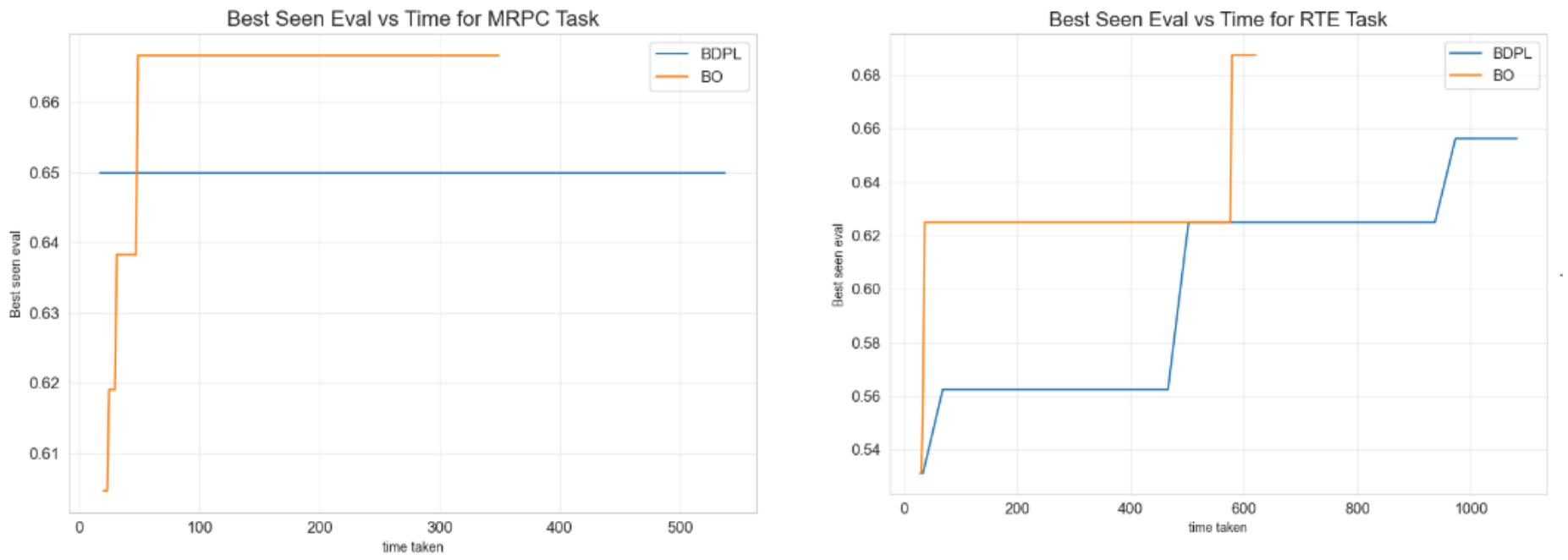

**Fig. 3.** Comparison of Best-Seen Performance for BO and BDPL over Time (in seconds)

**Table 4.** Execution timetable in seconds

| **Task** | **BO** | **BDPL** |
|---|---|---|
| MRPC | 379.02 | 571.75 |
| RTE | 661.92 | 1081.62 |
| QQP | 669.52 | 1070.14 |

This suggests two avenues for further investigation which account explicitly for the combinatorial nature of the search space.

- A specialized high dimensional version of BO.
- A better encoding than the naïve relaxation.

# 7 Analysis

## 7.1 Correctly predicted examples

| Task | **Prompt** + **Input** | Prediction | Label |
|---|---|---|---|
| MRPC | </s>This proposed investment starts the process of raising the funding we need to increase our capacity, " Elpida President Yukio Sakamoto said in a news release.</s>" The proposed investment starts the process of raising the funding we need to increase our capacity to better support our customer requirements," said Yukio Sakamoto, Elpida's president.?<mask>,</s> | semantically equivalent | Not semantically equivalent |
| | 20 steady especially Hollywood father contracts 130ined That abuse would fired water I key born positions saw 35 shares closed Rice inappropriate Sun missing Associates island cut technology warned Post feel planned training'Ministry Senate vetoed Sharon vulnerability aspects Ms noShe west first statements Republic dismissed regional </s>This proposed investment starts the process of raising the funding we need to increase our capacity, " Elpida President Yukio Sakamoto said in a news release.</s>" The proposed investment starts the process of raising the funding we need to increase our capacity to better support our customer requirements, " said Yukio Sakamoto, Elpida's president.?<mask>,</s> | Not semantically equivalent | |
| MRPC | </s>It also said it expects a civil complaint by the Securities and Exchange Commission.</s>Stewart also faces a separate investigation by the Securities and Exchange Commission.?<mask>,</s> | Not semantically equivalent | semantically equivalent |
| | 20 steady especially Hollywood father contracts 130ined That abuse would fired water I key born positions saw 35 shares closed Rice inappropriate Sun missing Associates island cut technology warned Post feel planned training'Ministry Senate vetoed Sharon vulnerability aspects Ms noShe west first statements Republic dismissed regional </s>It also said it expects a civil complaint by the Securities and Exchange Commission.</s>Stewart also faces a separate investigation by the Securities and Exchange Commission.?<mask>,</s> | semantically equivalent | |
| SST2 | <s>can you take before indigestion sets in  It was<mask>.</s> | great | terrible |
| | hook windowsallyoder 90 proficiency grandparentsreek metropolitanfredhy feathers 21 uses padding attention notes kids full flush moving debtsomem geout folksvideo rifeat fights famous strengths despair luc irre soft avoid racing black edge aliensrawn bug lob capable struggle di influenceieve <s>can you take before indigestion sets in  It was<mask>.</s> | terrible | |
| | <s> it 'll only put you to sleep.  It was<mask>.</s> | great | |
| SST2 | hook windowsallyoder 90 proficiency grandparentsreek metro-politanfredhy feathers 21 uses padding attention notes kids full flush moving debtsomem geout folksvideo rifeat fights famous strengths despair luc irre soft avoid racing black edge al-iensrawn bug lob capable struggle di influenceieve <s> it 'll only put you to sleep. It was<mask>.</s> | terrible | terrible |

**Table 5.** Here are four examples where the prompt obtained by BO made correct predictions.

### 7.2 Prompt length

**Table 6.** Score difference for different prompt length

| Task | Prompt length | Best score on test |
|---|---|---|
| Mrpc | 25 | 79.7619 |
| Mrpc | 50 | 78.4431 |
| Mrpc | 75 | 78.4195 |

To the impact of the prompt length on the performance it as not been given proper consideration: partial and preliminary results do not suggest that prompt length does not have a significant impact.

## 8 Conclusions and perspectives

The main conclusion of this paper is that Bayesian optimization is an effective tool for prompt optimization. The computational results obtained, even with a basic (vanilla) version of BO, compared with other approaches and “state of the art” results from the literature, for both white and black box prompting, confirm the feasibility of working directly in the token space using BO.

The key advantage of BO is that it does not need a categorical distribution over tokens nor a gradient and is suitable to PLM adaptation to down-stream tasks. This advantage gains traction also from the growing privacy and security issues of LLMs which are driving the adoption of MaaS (Models as a service) and related API based access.

A well-known limitation of Bayesian optimization is a degradation in its sample efficiency when the search space contains integer and categorical variables with conditional dependencies or graphs and strings: designing a kernel suitable to capture such structures has been receiving attention but still a widely agreed upon solution is not available. Another difficulty is related to the dimension of the search space, as is the case in combinatorial Bayesian optimization. For high dimensional BO several strategies have been proposed: three approaches which have been implemented in the Bo Torch library are quoted in sect. 4.

In this paper the natural language understanding tasks have been modelled as classification problems.

Furthermore, we have considered only a token-based search space for hard prompts which is of general application as it is the most general but is by no means unique. The versatility of BO should enable the extension of BO to alternative spaces as those proposed in (Zhang et al. , 2022) and (Prasad et al., 2023)

Another extension of BO as a prompting optimization tuning could focus on generative tasks which have been so far relatively less investigated. Also, the issues of prompt robustness (Maus et al., 2023) fairness are promising areas for developing context -specific Bayesian optimization methods.

**References.**

# 9 Appendix

## 9.1 A: BoTorch implementation

Bayesian optimization is performed using the BoTorch library built on PyTorch. A Gaussian process (GP) model is initialized to represent the objective function using the SingleTaskGP module. This models the objective as a GP with a single output for prompt performance. The ExactMarginalLogLikelihood module computes the exact log marginal likelihood for the GP posterior given the observations.

The acquisition function chosen for selecting the next prompt to evaluate is UpperConfidenceBound, which balances exploration and exploitation by maximizing the GP posterior mean plus β times the standard deviation.

The search space consists of possible prompt token indices, bounded between 0 and the maximum index normalized.

The prompt optimization task involves searching over possible sequences of discrete tokens to find the optimal prompt for a given task. When using a pre-trained masked language model like RoBERTa-large, the prompts must consist of tokens from its predefined vocabulary.

Specifically, the prompt is represented as a sequence of L discrete indices, with each index corresponding to one of the tokens in the vocabulary V. The vocabulary V contains |V| possible tokens, derived from the tokenization process during pre-training. For example, RoBERTa-large has a vocabulary with |V| = 50.265 tokens. In our implementation, each task has a specific candidate prompt vocabulary corresponding to a set of n-gram each representing a concatenation of RoBERTa-large model's tokens.

To construct the candidate prompt vocabulary, we use the script provided by (Diao et al. 2022 a) based on the code associated with the (Diao et al. 2021 b).

The search space can therefore be conceptualized as an L-dimensional discrete space, where L is the pre-defined prompt length for the given task. Each dimension ranges over the possible vocabulary indices {0, 1, ..., |V|-1}. Hence, the search space cardinality is |V|^L representing all possible prompt sequences of length L.

For example, in the case of mnli, the prompt length L=10 and |V|=117056, the cardinality of the prompt search space is 117.056*10.

To enable efficient optimization, we relax, for the Bayesian optimization process, discrete representation into a continuous space. Each discrete index is replaced with a continuous variable bounded between 0 and max index after normalization of the max index. Optimizing over this continuous relaxation, we round back to discrete indices (closest integer) to obtain the final prompt tokens.

This relaxation allows leveraging the power of Bayesian optimization over continuous spaces. The large discrete prompt search space is converted into a more tractable continuous optimization problem, while still maintaining a correspondence to discrete tokens through rounding. The continuous representation enables efficient exploration and exploitation over prompts using Gaussian process-based Bayesian optimization. The kernel used is Matern, ν=5/2.

### 9.2 Appendix B: Implementation Details

The experiments of Bayesian Optimization on RoBERTa-XL have been conducted on a machine instance with the following characteristics:

- CPU: 2 vCPUs @ 2.2 GHz
- RAM: 13 GB
- GPU: 1 x Tesla T4 GPU with 16 GB VRAM